\pdfoutput=1
\documentclass[conference]{IEEEtran}
\IEEEoverridecommandlockouts

\usepackage{cite}
\usepackage{amsmath,amssymb,amsfonts}  
\usepackage{algorithmic}
\usepackage{graphicx}
\usepackage{textcomp}                
\usepackage{xcolor}                   
\usepackage{booktabs}
\usepackage{url}
\usepackage{placeins}
\usepackage{hyperref}
\def\BibTeX{{\rm B\kern-.05em{\sc i\kern-.025em b}\kern-.08em
    T\kern-.1631em\lower.7ex\hbox{E}\kern-.125emX}}

\begin{document}

\title{A Multi-Model Approach to English-Bangla Sentiment Classification of Government Mobile Banking App Reviews}

\author{
\IEEEauthorblockN{\textbf{Md.~Naim Molla}\textsuperscript{1},
\textbf{Md~Muhtasim Munif Fahim}\textsuperscript{1},
\textbf{Md.~Binyamin}\textsuperscript{2},
\textbf{Md~Jahid Hasan Imran}\textsuperscript{1},\\
\textbf{Tonmoy Shil}\textsuperscript{1},
\textbf{Nura Rayhan}\textsuperscript{3},
and \textbf{Md~Rezaul Karim, PhD}\textsuperscript{1}\thanks{\protect\rule{0.95\columnwidth}{0.4pt}\protect\newline Corresponding author: Md~Rezaul Karim (mkarim@ru.ac.bd). Primary contact: Md.~Naim Molla (naim.molla.stats@gmail.com).}}
\IEEEauthorblockA{\textsuperscript{1}Data Science Research Lab, Department of Statistics, University of Rajshahi, Rajshahi-6205, Bangladesh}
\IEEEauthorblockA{\textsuperscript{2}Department of Statistics, Mawlana Bhashani Science and Technology University, Tangail-1902, Bangladesh}
\IEEEauthorblockA{\textsuperscript{3}Department of Statistics, University of Rajshahi, Rajshahi-6205, Bangladesh}
}

\maketitle

\begin{abstract}
For millions of users in developing economies who depend on mobile banking as their primary gateway to financial services, app quality directly shapes financial access. The study analyzed 5,652 Google Play reviews in English and Bangla (filtered from 11,414 raw reviews) for four Bangladeshi government banking apps. The authors used a hybrid labeling approach that combined use of the reviewer's star rating for each review along with a separate independent XLM-RoBERTa classifier to produce moderate inter-method agreement ($\kappa = 0.459$). Traditional models outperformed transformer-based ones: Random Forest produced the highest accuracy (0.815), while Linear SVM produced the highest weighted F1 score (0.804); both were higher than the performance of fine-tuned XLM-RoBERTa (0.793). McNemar's test confirmed that all classical models were significantly superior to the off-the-shelf XLM-RoBERTa ($p < 0.05$), while differences with the fine-tuned variant were not statistically significant. DeBERTa-v3 was applied to analyze the sentiment at the aspect level across the reviews for the four apps; the reviewers expressed their dissatisfaction primarily with the speed of transactions and with the poor design of interfaces; eJanata app received the worst ratings from the reviewers across all apps. Three policy recommendations are made based on these findings—remediation of app quality, trust-centred release management, and Bangla-first NLP adoption—to assist state-owned banks in moving towards improving their digital services through data-driven methods. Notably, a 16.1-percentage-point accuracy gap between Bangla and English text highlights the need for low-resource language model development.
\end{abstract}


\section{Introduction}

The use of mobile banking is increasing at an incredibly fast pace in Bangladesh as many customers of the four main state-owned banks (Sonali Bank, Agrani Bank, Janata Bank, Rupali Bank) have downloaded the banking applications to the Google Play Store and now have direct access to their bank accounts through this means~\cite{b1}. These applications are particularly important for the large number of government employees, retirees and people living in rural areas who do not have the same level of access to private-sector banking as those in urban areas~\cite{b2}. Reviews of users on the Google Play Store are a free source of information that can be used to determine how satisfied users are with the services they receive from the banks; however, they may also identify service failures and other issues~\cite{b3}.

The process of using automated methods to analyze user reviews of mobile banking applications poses several problems. First, user reviews of mobile banking applications often contain multiple languages including English, Bangla script and Romanized Bangla. As such, the effectiveness of most monolingual pipelines will depend upon the ability to translate the review into a language that can be processed by the pipeline; however, this can introduce additional error or ``noise''~\cite{b4}. Second, while there have been studies on the use of natural language processing (NLP) to analyze user reviews of mobile banking applications, to date, there has been no study that has analyzed user reviews of all four of the state-owned banking applications using a unified bilingual corpus and comparing the performance of both classical and transformer-based machine learning models to classify the sentiment expressed in the reviews. Much of the existing research in Bangla NLP related to fintech has focused on monolingual non-banking corpora~\cite{b5,b6}.

This paper makes the following contributions:
\begin{enumerate}
    \item A bilingual (English--Bangla) sentiment dataset created specifically for the purpose of evaluating user experience with the mobile banking applications of the four state-owned banks in Bangladesh, which was developed using a hybrid star-rating/model-validated labeling approach.
    \item A systematic comparison of four classical machine learning models (Na\"ive Bayes, Linear SVM, Logistic Regression, Random Forest) against off-the-shelf (OTS) and fine-tuned XLM-RoBERTa~\cite{b4}, with McNemar's test~\cite{b8} and 95\% bootstrap confidence intervals to evaluate statistical significance.
    \item Aspect-level sentiment analysis using DeBERTa-v3~\cite{b9} to evaluate the degree to which users were expressing positive or negative sentiment towards each of six service dimensions.
    \item An evaluation of the degree to which there is a difference in performance between the English and Bangla reviews, and an examination of the longitudinal trend in sentiment expressed by users of the mobile banking applications of the four state-owned banks in Bangladesh from 2021--2025.
    \item Actionable policy recommendations for state-owned banks based on the sentiment findings, including performance service level agreements (SLAs), trust-based release management, and the implementation of Bangla-first NLP pipelines to address linguistic equity.
\end{enumerate}

\section{Related Work}

\textbf{Sentiment and NLP of app reviews.}
One of the biggest challenges with collecting data for app reviews is the gap between what users rate their apps (e.g., 1 or 5) and the written comments they leave about their experience with an app~\cite{b10,b11}. There has been a number of very successful models using BERT~\cite{b12} and RoBERTa~\cite{b13} on a wide variety of English review datasets~\cite{b14}. However, Bangla, which also has many of these same characteristics (agglutinative morphology, frequent use of code-switching, etc.), presents a much greater challenge. While there are some notable examples of monolingual and multilingual efforts in this space (i.e., SentNoB~\cite{b5}, BanglaBook~\cite{b15}, and BanglaBERT~\cite{b16}), no team participating in the BLP-2023 shared task (71 total teams)~\cite{b17} focused on the banking domain.

\textbf{Multilingual transformers and ABSA.}
The fact that the multilingual version of RoBERTa (XLM-RoBERTa~\cite{b4}) was trained on over 100 languages makes it easier to leverage knowledge gained from one language to improve model performance on other languages. This includes leveraging the Twitter version of XLM-RoBERTa (XLM-T~\cite{b7}) and additional prompt-tuning variations~\cite{b18} to perform zero-shot polarity classification. Although the SemEval-2014~\cite{b20} remains the standard benchmark for ABSA research, recent results show that DeBERTa~\cite{b9}'s attention mechanism (which is specifically designed to be able to disentangle different levels of abstraction) can achieve top results on several ABSA benchmark tasks~\cite{b21}, and that additional cross-lingual alignment techniques~\cite{b22} and hybrid architectures (such as Instruct-DeBERTa~\cite{b23}) will likely continue to break new ground in the field. Yet, we could find no prior research that performs bilingual financial-service evaluation.

\textbf{Sentiment analysis of mobile banking apps.}
Prior research into banking app user feedback frequently focuses on three main aspects: (1) how easy it is to use the app, (2) whether the app appears trustworthy to the user, and (3) how reliable the app is to avoid crashing~\cite{b24}. Prior research has also demonstrated the ability to identify common latent topics associated with banking app reviews using topic modelling approaches such as BERTopic and LDA~\cite{b25}. Additionally, prior work demonstrates that negative spikes in user reviews are most commonly found after software updates~\cite{b26,b27}. Published research in mobile financial services within Bangladesh almost exclusively focuses on private mobile financial services providers (e.g., bKash), and thus does not examine the sentiment analysis of mobile banking apps provided by the state-owned banks, nor does it investigate the use of aspect-level transformer analysis~\cite{b28}.

\section{Dataset}

We used \texttt{google-play-scraper} to extract the reviews for Sonali e-Wallet, Agrani Smart, eJanata and Rupali e-Bank on the Google Play Store. The corpus covers reviews posted between January 2021 and December 2025, collected in January 2026. Each row of data includes the text of the review, the star rating assigned to the review (1--5), the time and date of when it was posted, how many users have liked or thumbs-upped the review (\textit{thumbsUpCount}), and the version of the app at the time the review was made. We removed duplicate records and noisy records, and after this removal process we retained 7,044 out of 11,414 records (a retention rate of 61.7\%). Language detection was then performed using \texttt{langdetect}; 1,379 reviews (19.6\%) not classified as English or Bangla were removed, yielding a bilingual corpus of 5,665 reviews (English: 4,540, 80.1\%; Bangla: 1,125, 19.9\%). After text preprocessing (lowercasing, URL and emoji removal, stop-word filtering), 13 empty reviews were discarded, leaving 5,652 reviews for analysis. A summary of the initial cleaning stage is contained in Table~\ref{tab:dataset_stats}.

\begin{table}[htbp]
\caption{Dataset Statistics per Application}
\label{tab:dataset_stats}
\begin{center}
\begin{tabular}{l c c c}
\toprule
\textbf{App} & \textbf{Raw} & \textbf{Clean} & \textbf{Avg.\ Rating} \\
\midrule
Sonali e-Wallet & 7,306 & 4,347 & 3.81 \\
Agrani Smart & 2,434 & 1,432 & 3.27 \\
eJanata & 1,232 & 925 & 2.55 \\
Rupali e-Bank & 442 & 340 & 3.49 \\
\midrule
\textbf{Total} & \textbf{11,414} & \textbf{7,044} & \textbf{3.52} \\
\bottomrule
\end{tabular}
\end{center}
\end{table}

\section{Methodology}

\begin{figure*}[t!]
\centerline{\includegraphics[width=\textwidth]{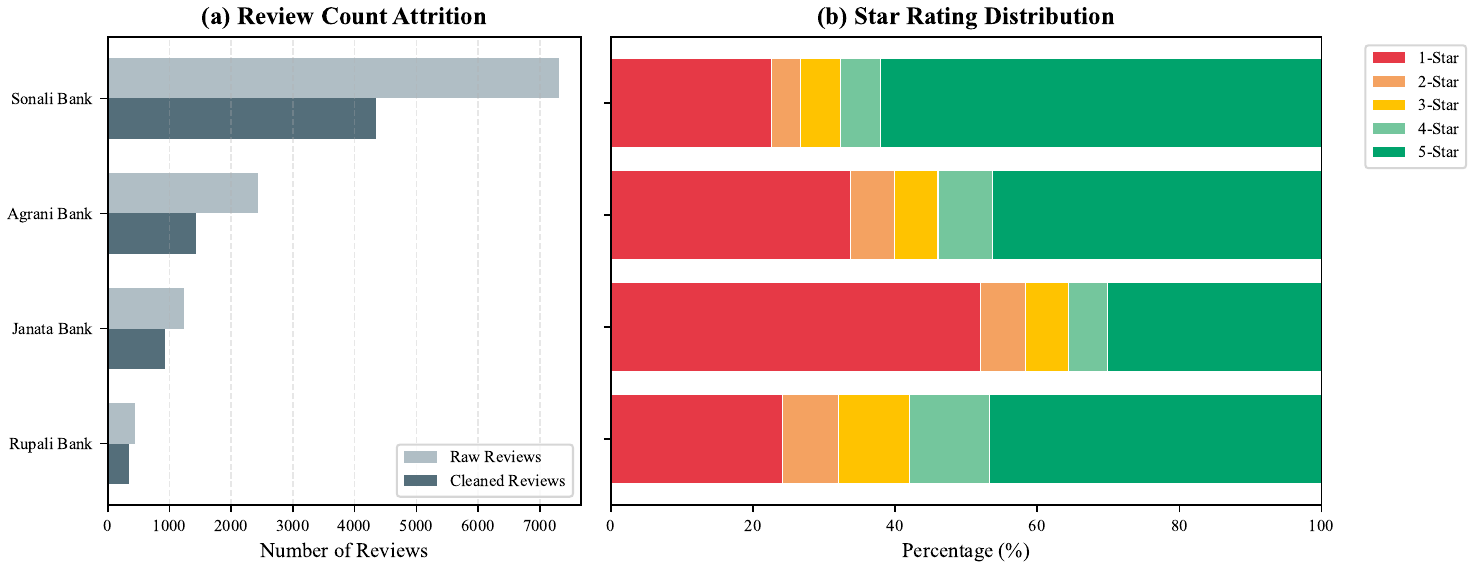}}
\caption{Distribution of raw and cleaned reviews across the four applications.}
\label{fig:dataset_stats}
\end{figure*}

\subsection{Pipeline Overview}

The bilingual corpus of 5,652 reviews was divided 80/20 into 4,521 training examples and 1,131 held-out test examples, with stratification of classes. The held-out set is labeled using star-rating-based sentiment only, while the training set undergoes additional consensus filtering (\S\ref{sec:relabel}). Three parallel classification tracks are used: OTS XLM-RoBERTa, fine-tuned XLM-RoBERTa, and four TF-IDF-based classical classifiers. In addition, a separate DeBERTa-v3 ABSA pipeline is used to generate aspect-level sentiment profiles. All experiments were performed on a single NVIDIA RTX 3060 (with 12\,GB VRAM).

\subsection{Hybrid Relabeling}\label{sec:relabel}

Firstly, the star ratings are taken as a proxy to determine whether a review has a negative or positive sentiment: 1--2 stars $\to$ negative, 3 $\to$ neutral, 4--5 stars $\to$ positive. Secondly, to minimize label noise, every review is classified by the \texttt{cardiffnlp/}\allowbreak\texttt{twitter-xlm-}\allowbreak\texttt{roberta-}\allowbreak\texttt{base-}\allowbreak\texttt{sentiment} model. If the star-based label does not match the model prediction, the sample is removed from the training data. All 4,521 training samples were classified by both the star-rating heuristic and the model; 1,564 (34.6\%) were inconsistent and removed, resulting in 2,957 consensus-labeled training instances (English: 2,521, 85.3\%; Bangla: 436, 14.7\%). Inter-method agreement is quantified by Cohen's $\kappa$:
\begin{equation}
\kappa = \frac{p_o - p_e}{1 - p_e}
\end{equation}
where $p_o$ is observed agreement and $p_e$ is chance agreement. We obtain $\kappa = 0.459$ (moderate agreement).

\subsection{XLM-RoBERTa Classification}

The \texttt{cardiffnlp/}\allowbreak\texttt{twitter-xlm-}\allowbreak\texttt{roberta-}\allowbreak\texttt{base-}\allowbreak\texttt{sentiment} model, that has been pretrained on 2.5\,B multilingual social-media tokens, is evaluated in two configurations: (i)~OTS, which is not task-specifically trained; and (ii)~fine-tuned for 3 epochs (AdamW, learning rate $= 2 \times 10^{-5}$, batch size $= 16$, max length $= 128$) on a stratified subsample of 1,200 reviews drawn from the 2,957 consensus training set (balanced across 3 sentiment classes $\times$ 2 languages) using 5-fold cross-validation.

\subsection{Classical Baselines}

Four scikit-learn~\cite{b29} classifiers---Na\"ive Bayes, Linear SVM, Logistic Regression, and Random Forest---are trained on TF-IDF features (unigrams + bigrams, maximum 15,000 features, sublinear TF). Hyperparameters are determined by GridSearchCV (5-fold, macro-F1 objective). All models are evaluated on the same 1,131-sample held-out test set.

\subsection{Aspect-Based Sentiment Analysis}

We use \texttt{yangheng/deberta-v3-base-absa-v1.1}, fine-tuned on SemEval ABSA data. Every review is paired with six aspect terms: UI/UX, Security, Speed/Performance, Customer Service, Features, and Transaction Processing. The model returns a polarity label and confidence score per pair; reviews without detectable aspect cues are excluded. The results are aggregated per application.

\section{Results}

\begin{figure*}[t!]
\centerline{\includegraphics[width=\textwidth]{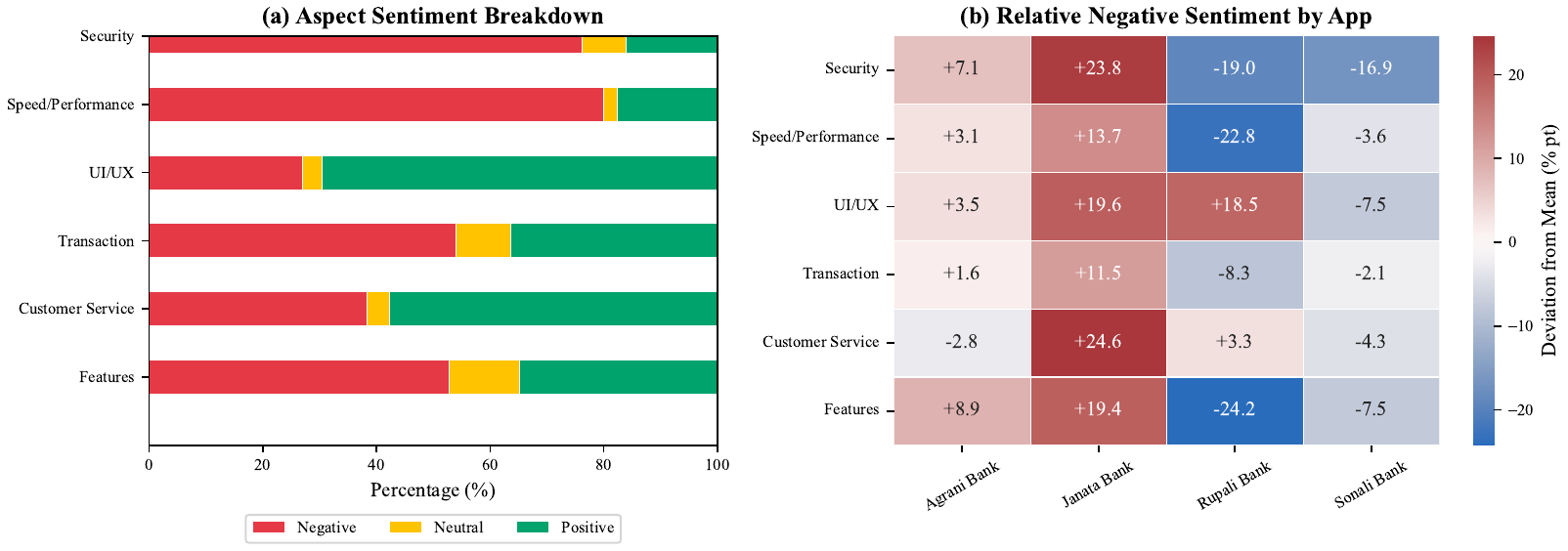}}
\caption{Aspect-based sentiment polarity across six service dimensions.}
\label{fig:absa_results}
\end{figure*}

\subsection{Model Comparison}

Accuracy, precision, recall, weighted F1, and the 95\% bootstrap confidence intervals of the results are presented in Table~\ref{tab:model_perf}. Two immediate observations can be made about the table. Firstly, Random Forest is the most accurate model with an accuracy of 0.815 (CI [0.778, 0.829]). Linear SVM has the highest weighted F1 score of 0.804. Both classical learning algorithms have a higher accuracy and weighted F1 score than the fine-tuned XLM-RoBERTa (0.793), while the off-the-shelf variant achieves only 0.740. Secondly, the difference between the top classical algorithm and the fine-tuned transformer is relatively small. We attribute this narrow gap to the limited size of the consensus training set ($n = 2{,}957$), which is insufficient for a large transformer model to fully adapt to the domain-specific vocabulary and sentiment patterns of bilingual banking reviews.

\begin{table}[htbp]
\caption{Model Performance on Test Set (Weighted F1)}
\label{tab:model_perf}
\begin{center}
\begin{tabular}{l c c c c c}
\toprule
\textbf{Model} & \textbf{Acc.} & \textbf{Prec.} & \textbf{Rec.} & \textbf{W-F1} & \textbf{95\% CI} \\
\midrule
Random Forest & 0.815 & 0.801 & 0.815 & 0.803 & [.778, .829] \\
Linear SVM & 0.809 & 0.808 & 0.809 & 0.804 & [.781, .829] \\
Log. Reg. & 0.804 & 0.801 & 0.804 & 0.798 & [.773, .823] \\
Na\"ive Bayes & 0.814 & 0.801 & 0.814 & 0.795 & [.772, .821] \\
XLM-R (FT) & 0.769 & 0.834 & 0.769 & 0.793 & [.772, .818] \\
XLM-R (OTS) & 0.683 & 0.839 & 0.683 & 0.740 & [.716, .764] \\
\bottomrule
\end{tabular}
\end{center}
\end{table}

\subsection{Statistical Significance}

As shown in McNemar's test~\cite{b8} on the held-out predictions, all classical models perform better than the OTS XLM-RoBERTa ($p < 0.001$; Table~\ref{tab:mcnemar}). Furthermore, the differences between classical models and the fine-tuned XLM-RoBERTa are very small ($\Delta$\,W-F1 $\leq 0.011$) and thus not statistically significant, indicating that the performance of both types of models are similar at the current sample size.

\begin{table}[htbp]
\caption{McNemar's Test: OTS XLM-RoBERTa vs.\ Classical Models}
\label{tab:mcnemar}
\begin{center}
\begin{tabular}{l c c c}
\toprule
\textbf{Comparison} & $\chi^2$ & \textbf{p-value} & \textbf{Sig.} \\
\midrule
vs.\ Logistic Reg. & 76.11 & $<$0.001 & Yes \\
vs.\ Linear SVM & 82.97 & $<$0.001 & Yes \\
vs.\ Random Forest & 80.82 & $<$0.001 & Yes \\
vs.\ Na\"ive Bayes & 81.85 & $<$0.001 & Yes \\
\bottomrule
\end{tabular}
\end{center}
\end{table}

\subsection{Cross-Application Sentiment Rankings}

Using the 2,957 consensus-labeled reviews, the \textit{thumbsUpCount}-weighted Positive Sentiment Score (PSS) and Negative Sentiment Score (NSS) are computed per app (reviews with zero \textit{thumbsUpCount} receive zero weight):
\begin{equation}
\text{PSS}(a) = \frac{\sum_{i \in \mathcal{P}_a} v_i}{\sum_{i \in \mathcal{R}_a} v_i} \times 100\%
\end{equation}
where $\mathcal{P}_a$ is the set of positive reviews for app $a$, $\mathcal{R}_a$ the full review set, and $v_i$ the \textit{thumbsUpCount} of review $i$. NSS is defined analogously over negative reviews.

As presented in Table~\ref{tab:rankings}, Rupali e-Bank is ranked first (PSS\,=\,58.4\%), followed by Sonali e-Wallet (52.8\%). Agrani Smart has a high negative sentiment score (NSS\,=\,66.6\%), whereas eJanata is ranked last with the highest NSS (80.4\%) and lowest average rating (mean\,=\,2.20).

\begin{table}[htbp]
\caption{Cross-Application Sentiment Rankings (ThumbsUp-Weighted)}
\label{tab:rankings}
\begin{center}
\begin{tabular}{l c c c c}
\toprule
\textbf{Application} & \textbf{PSS} & \textbf{NSS} & \textbf{Neutral} & \textbf{Avg.} \\
& (\%) & (\%) & (\%) & \textbf{Rating} \\
\midrule
Rupali e-Bank & 58.4 & 34.5 & 7.07 & 3.50 \\
Sonali e-Wallet & 52.8 & 41.8 & 5.32 & 3.67 \\
Agrani Smart & 29.1 & 66.6 & 4.19 & 3.06 \\
eJanata & 16.3 & 80.4 & 3.15 & 2.20 \\
\bottomrule
\end{tabular}
\end{center}
\end{table}

\subsection{Aspect-Based Sentiment Analysis}

According to the DeBERTa-v3 ABSA model, the primary cause of negative sentiment across all apps is the Speed/Performance aspect. eJanata is severely affected, with 61.3\% of its speed-related mentions being negatively labeled compared to 35.2\% for Sonali e-Wallet. The second major contributor to negative sentiment is the UI/UX interface quality, where 52.4\% of eJanata's UI/UX-related mentions are negative, whereas it is 21.3\% for Sonali. Interestingly, although security-related complaints surface in only 18.7--31.8\% of aspect mentions, they accumulate substantially higher \textit{thumbsUp} counts than other categories, suggesting that such concerns resonate well beyond the individual reviewer. Fig.~\ref{fig:absa_results} displays the detailed distribution of the sentiment across all aspects.

\begin{figure*}[t!]
\centerline{\includegraphics[width=\textwidth]{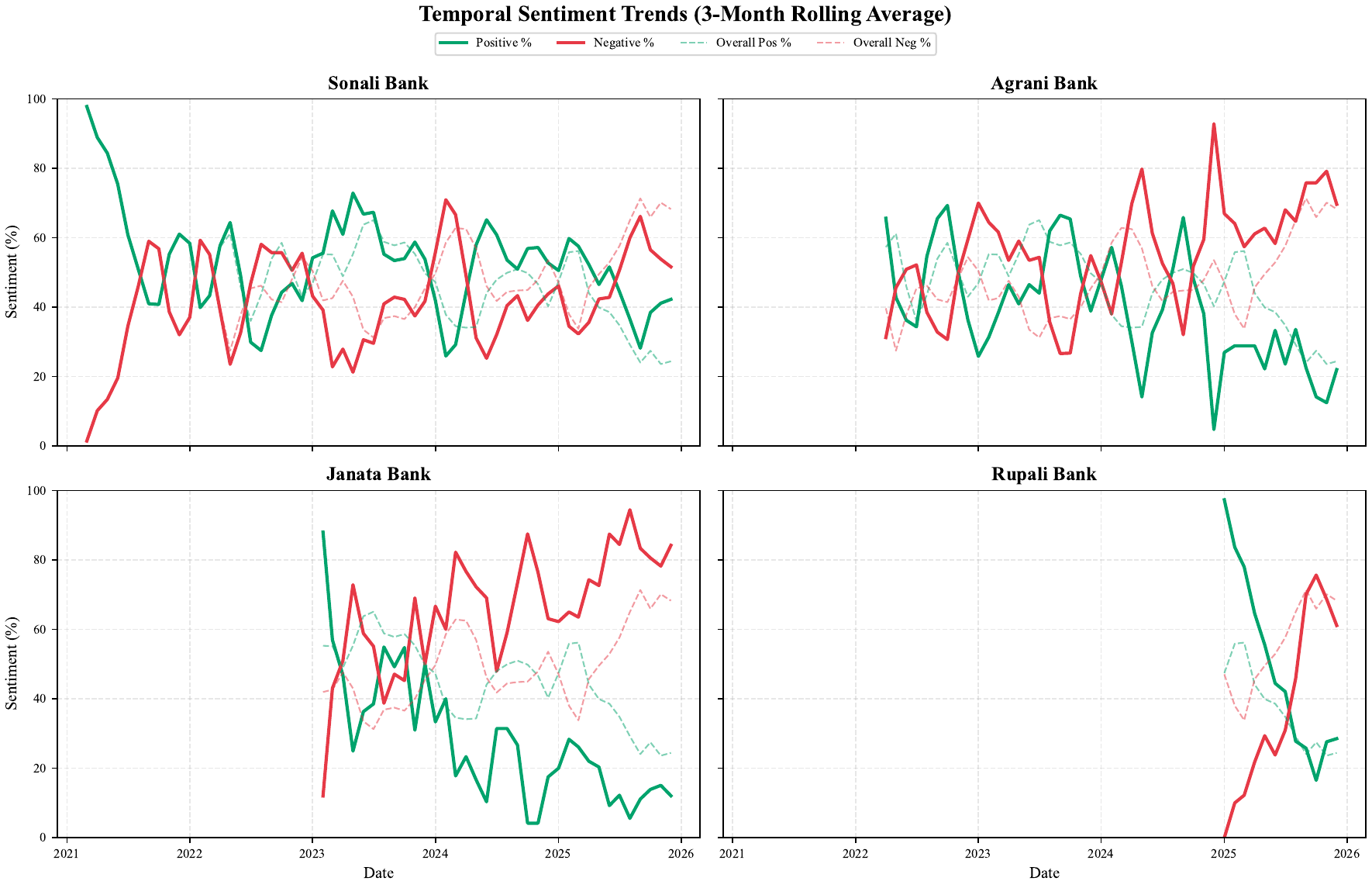}}
\caption{Longitudinal sentiment trends (2021--2025) across the four applications.}
\label{fig:temporal_trends}
\end{figure*}

\subsection{Language-Stratified Evaluation}

When the test set is split based on the detected script language (Table~\ref{tab:lang_eval}), the results show a disturbing picture. On English reviews, the fine-tuned XLM-RoBERTa achieves an accuracy of 0.715, whereas on Bangla reviews, the accuracy drops to 0.554---a difference of 16.1 percentage points. The same holds true for the weighted F1 scores, which also drop from 0.763 to 0.633. These asymmetries are caused by three interconnected factors. First, the pretraining corpus used for the model consists mainly of English web texts. Second, Bangla tokens are fragmented into finer subwords than English tokens, reducing the semantic coherence within the tokenized texts. Third, Bangla orthography is characterized by a great deal of variability, especially when using informal orthography, resulting in users frequently switching between different scripts or even creating their own spelling rules. In addition to the model-based accuracy of the sentiment analysis, these asymmetries represent an equity issue: users writing in Bangla---often representing less digitally connected communities---would systematically receive lower-quality sentiment labels in any deployed triage system.

\begin{table}[htbp]
\caption{Language-Stratified Evaluation (Fine-Tuned XLM-RoBERTa)}
\label{tab:lang_eval}
\begin{center}
\begin{tabular}{l c c c c}
\toprule
\textbf{Language} & \textbf{Acc.} & \textbf{W-F1} & \textbf{M-F1} & \textbf{F1 CI} \\
\midrule
English & 0.715 & 0.763 & 0.599 & [.738, .790] \\
Bangla & 0.554 & 0.633 & 0.499 & [.564, .692] \\
\midrule
\textbf{Gap} & 0.161 & 0.130 & 0.100 & --- \\
\bottomrule
\end{tabular}
\end{center}
\end{table}

\subsection{Temporal Trends}

The monthly sentiment proportions from 2021 through 2025 show an increase of 17 percentage points in negative polarity. There is a clear relationship between negativity peaks and app updates. Post-update patches lead to temporary improvements; however, the long-term trend of increasing negativity is evident. eJanata clearly declines the most, while Sonali e-Wallet appears to be somewhat stable (Fig.~\ref{fig:temporal_trends}).

\subsection{Recommendations for State-Owned Banks in Bangladesh Based on Study Findings}

Based on this study's findings, there are three actionable policy recommendations that can be made for state-owned banks in Bangladesh:

\textbf{(1)~Remediate app quality issues related to performance/speed and User Experience/Usability.} The data from the study show that 61.3\% of eJanata's negative sentiment is due to performance/speed while 52.4\% is due to User Experience/Usability. These percentages are nearly twice those of Sonali e-Wallet. State-owned banks should develop service level agreements (SLA) for their apps' performance and test them using bilingual usability tests prior to major releases.

\textbf{(2)~Implement trust-based release management.} The data from the study shows that negative sentiment spikes have consistently been associated with app releases but that post-release patches were only able to reduce the spike in negativity by a maximum of 17 percentage points for the period studied. Additionally, although security complaints were relatively rare (only 18.7--31.8\% of all aspect mentions) they generated disproportionately high ``thumbs up'' counts. To mitigate these types of spikes and maintain user trust, state-owned banks should use staged rollout strategies incorporating beta-testing cohorts as well as provide users with real-time access to sentiment dashboards; furthermore, state-owned banks should proactively disclose results of their security audits prior to app releases.

\textbf{(3)~Use Bangla-first Natural Language Processing (NLP) to enable equitable complaint routing based upon language.} The 16.1-percentage-point accuracy difference between Bangla and English reviews indicates that if an automated complaint-routing system was used, it would systematically under-serve Bangla-speaking users, who are often from rural or less digitally connected communities. As such, both state-owned banks and regulatory agencies should require that domain-adapted Bangla models (for example, BanglaBERT~\cite{b16}) be used in customer feedback processing pipelines to promote linguistic equity in state-owned bank service delivery.

\subsection{Limitations}

Our findings have several limitations. First, the Google Play corpus excludes feature-phone USSD and agent-banking users. Second, mapping three-star ratings to a neutral category introduces subjective noise ($\kappa = 0.459$). Third, the consensus-filtered training set is heavily skewed toward English (85.3\%), which, alongside a distributional mismatch in the unfiltered test set, may disproportionately inhibit Bangla classification performance. Finally, the English-only pretraining of the DeBERTa-v3 model limits its ability to robustly extract Bangla aspect cues, though the lack of domain-specific bilingual ABSA benchmarks hinders precise error calibration.

\section{Conclusion}

We have built a bilingual corpus of 5,652 English and Bangla Google Play reviews (filtered from 11,414 raw reviews) from four Bangladeshi public-sector banking apps and compared the performance of six sentiment analysis models. The classical models (Random Forest: W-F1\,=\,0.803; Linear SVM: W-F1\,=\,0.804) performed better than the fine-tuned version of XLM-RoBERTa (W-F1\,=\,0.793) on this dataset and McNemar's test confirmed that they were significantly superior to the OTS version ($p < 0.05$). DeBERTa-v3 ABSA identified speed and UI/UX as the two main complaint categories and showed that eJanata was the worst performer across all metrics. The 16.1-percentage-point difference in accuracy of the English and Bangla reviews shows that there is a clear requirement for domain-adapted Bangla models such as BanglaBERT~\cite{b16} to improve performance. The next phase of our research will include extending the scope to cover private-sector banking apps and exploring whether domain-specific pretraining can be used to address the cross-lingual performance gap.

\FloatBarrier

\end{document}